\documentclass[10pt,twocolumn,letterpaper]{article}

\usepackage[pagenumbers]{cvpr} % To force page numbers, e.g. for an arXiv version

\definecolor{cvprblue}{rgb}{0.21,0.49,0.74}
\usepackage[pagebackref,breaklinks,colorlinks,allcolors=cvprblue]{hyperref}
\usepackage{rotating}
\usepackage{graphicx}  % \resizebox、\rotatebox
\usepackage{rotating}  % sideways 环境
\usepackage{xcolor}    % \definecolor、\textcolor
\usepackage{booktabs}  % \toprule、\midrule、\bottomrule
\usepackage{pifont}    % \ding
\usepackage{amssymb}   % \blacksquare
\usepackage{multirow}  % multirow
\usepackage{soul}
\usepackage[ruled,vlined]{algorithm2e}
\def\confName{CVPR}
\def\confYear{2027}

\title{Seeing 3D Occupancy Through Scene-Adaptive Projection}

\author{
Yuan Wu$^{1}$ \quad
Zhiqiang Yan$^{2}$\footnotemark[1] \quad
Zhengxue Wang$^{1}$ \quad
Guangfa Wang$^{1}$ \quad
Jun Li$^{1}$ \quad
Jian Yang$^{1}$\footnotemark[1]\\
$^{1}$Nanjing University of Science and Technology \quad
$^{2}$National University of Singapore\\
}

\newcommand{\first}[1]{\textbf{#1}}
\newcommand{\second}[1]{\underline{#1}}
\definecolor{barriercolor}{RGB}{255, 120, 50}
\definecolor{bicyclecolor}{RGB}{255, 192, 203}
\definecolor{buscolor}{RGB}{255, 255, 0}
\definecolor{carcolor}{RGB}{0, 150, 245}
\definecolor{constructcolor}{RGB}{0, 255, 255}
\definecolor{motorcolor}{RGB}{200, 180, 0}
\definecolor{pedestriancolor}{RGB}{255, 0, 0}
\definecolor{trafficcolor}{RGB}{255, 240, 150}
\definecolor{trailercolor}{RGB}{135, 60, 0}
\definecolor{truckcolor}{RGB}{160, 32, 240}
\definecolor{drivablecolor}{RGB}{255, 0, 255}
\definecolor{otherflatcolor}{RGB}{139, 137, 137}
\definecolor{sidewalkcolor}{RGB}{75, 0, 75}
\definecolor{terraincolor}{RGB}{150, 240, 80}
\definecolor{manmadecolor}{RGB}{230, 230, 255}
\definecolor{vegetationcolor}{RGB}{0, 175, 0}
\definecolor{otherscolor}{RGB}{0, 0, 0}
\begin{document}
\maketitle
\footnotetext[1]{Corresponding authors.}
\begin{abstract}
Vision-based 3D occupancy prediction infers voxel-wise geometry and semantics from multi-view images, where 2D-to-3D view transformation is a crucial step.
Many existing methods adopt a backward-projection paradigm, projecting uniformly distributed 3D reference points onto image planes to aggregate multi-view features.
However, their scene-agnostic sampling patterns overlook the non-uniform distribution of scene geometry, allocating many references to empty space and reducing sampling efficiency under a fixed sampling budget.
To address this challenge, we propose \textbf{SAP}, a \textbf{s}cene-\textbf{a}daptive \textbf{p}rojection framework that leverages depth-based geometric priors to guide view transformation.
Specifically, we back-project metric depth maps predicted by a frozen depth model into 3D space to obtain geometric priors.
To leverage these priors, we design a Scene-Adaptive Projection Encoder comprising two complementary components.
First, Prior-Guided Reference Generation constructs a hybrid reference set for feature sampling, allocating references within prior-supported height regions and retaining full-range sampling where valid support is unavailable.
Furthermore, to account for potential spatial misalignment caused by depth estimation errors, Radial-Structured Feature Aggregation explores multiple candidate locations through learned radial offsets to aggregate potentially displaced feature evidence. 
Extensive experiments on Occ3D demonstrate that SAP achieves state-of-the-art performance and consistently improves results across diverse occupancy frameworks.
%
% Code will be released.
\end{abstract}
    
\section{Introduction}
\label{sec:intro}

3D occupancy prediction aims to infer voxel-wise geometry and semantics of the surrounding environment from multi-sensor observations, providing a unified scene representation for holistic perception and motion planning in autonomous driving~\cite{du2026sparseworld,liu2026drivepi,yang2025driving,zheng2024occworld,lian2026liauto}. Among different sensing modalities, vision-based occupancy prediction has attracted growing attention due to the rich semantic information and low deployment cost of cameras. 
%%%%%%%%%%%%%%%%%%%%%%%%%%%%%%%%%%%%%%%%%
\begin{figure}[t]
    \centering
    \includegraphics[width=1.0\columnwidth]{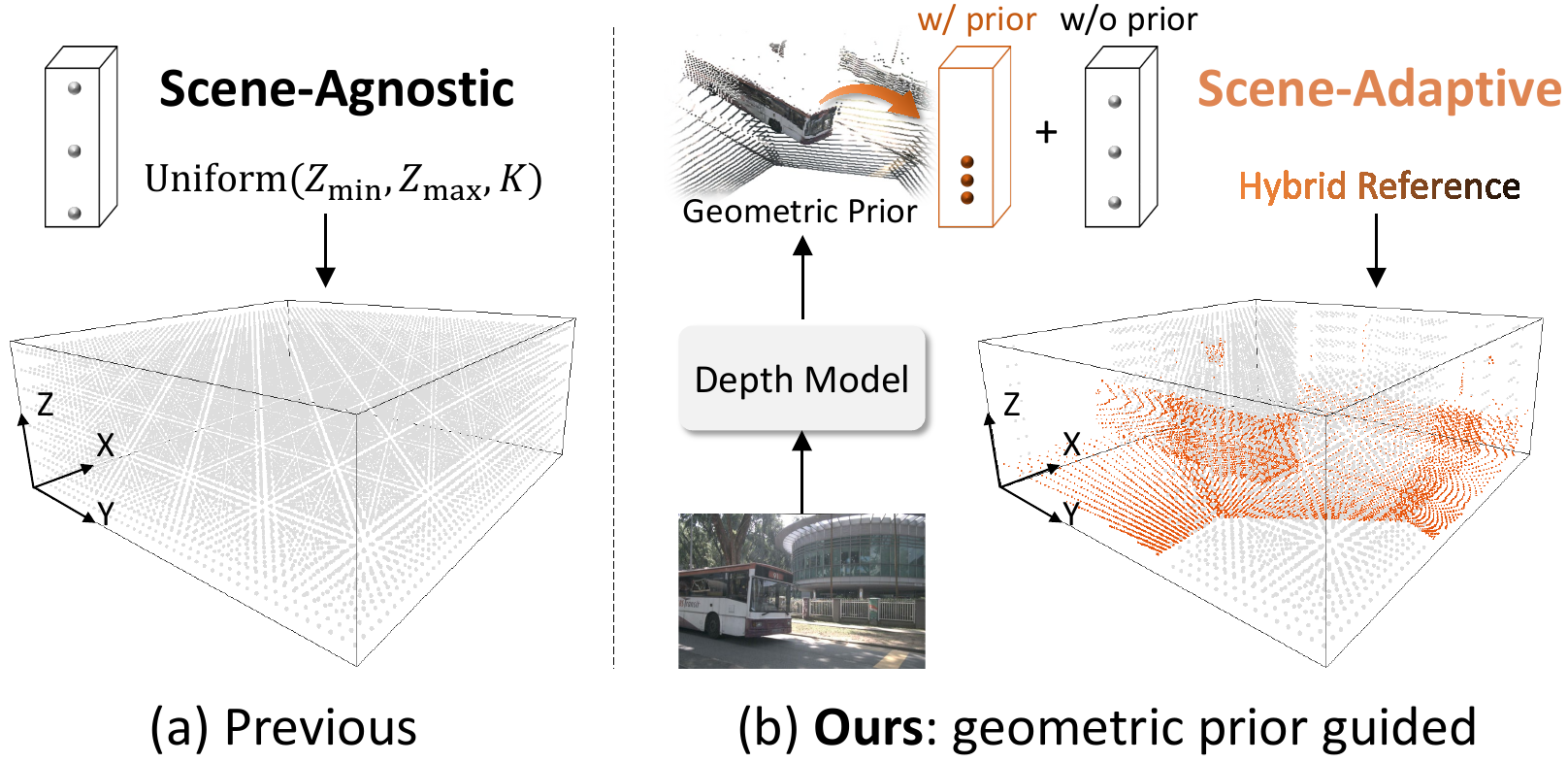}
    \vspace{-14pt}
    \caption{Comparison of reference sampling strategies.
    (a) Previous methods adopt scene-agnostic sampling patterns, uniformly distributing references regardless of scene geometry.
    (b) Ours leverages depth-based geometric priors to generate scene-adaptive hybrid references that prioritize geometrically supported regions.}
    \label{fig:teaser}
    \vspace{-16pt}
\end{figure}
%%%%%%%%%%%%%%%%%%%%%%%%%%%%%%%%%%%%%%%%%

Despite recent advances, effectively transforming 2D image features into 3D space remains a critical challenge. Many existing methods~\cite{li2024bevformer,li2023fb,liao2025stcocc,wu2026see} follow a backward-projection paradigm, where predefined 3D reference points are projected onto image planes to aggregate multi-view features. As illustrated in Fig.~\ref{fig:teaser}(a), these methods typically adopt scene-agnostic reference patterns, applying the same sampling strategy across BEV locations regardless of the underlying scene geometry. However, real-world scenes exhibit highly non-uniform geometric distributions, with relevant structures occupying only a fraction of the predefined sampling space. 
Consequently, many reference points may fall into regions with limited geometric support, reducing sampling efficiency under a fixed reference budget.

To address this challenge, we propose SAP, a scene-adaptive projection framework that leverages depth-based geometric priors to guide 2D-to-3D view transformation. 
Specifically, we employ a frozen depth prediction model to estimate metric depth maps, which are back-projected into 3D space to obtain geometric priors reflecting the spatial distribution of scene structures. These priors provide scene-adaptive geometric guidance for reference sampling. Based on this, we introduce a Scene-Adaptive Projection Encoder comprising two complementary components.

We first introduce Prior-Guided Reference Generation (PGRG), which leverages geometric priors to construct a hybrid reference set. Specifically, PGRG identifies supported height intervals within each BEV pillar, preserving potentially discontinuous and multi-layer vertical structures. For pillars with valid prior support, references are allocated within these intervals, whereas full-range sampling is retained for BEV pillars without valid support. 
As illustrated in Fig.~\ref{fig:teaser}(b), the resulting hybrid reference set concentrates references in geometrically supported regions rather than empty space, enabling scene-adaptive sampling.

While geometric priors provide valuable guidance, depth estimation errors can displace back-projected points along camera rays, introducing approximately radial spatial misalignment in BEV space. To account for this potential misalignment, we further introduce Radial-Structured Feature Aggregation (RSFA), which enables each BEV query to explore multiple candidate locations through learned radial offsets and aggregate potentially displaced BEV features.

In summary, our main contributions are as follows:
\begin{itemize}
    \item
    We propose SAP, a scene-adaptive projection framework that leverages depth-based geometric priors to guide 2D-to-3D view transformation.

    \item We design a Scene-Adaptive Projection Encoder: PGRG exploits geometric priors to construct a scene-adaptive hybrid reference set, while RSFA aggregates potentially misaligned features through learned radial offsets.

    \item
    Extensive experiments demonstrate that SAP achieves state-of-the-art performance on Occ3D and consistently improves results across different occupancy models.
\end{itemize}
\section{Related Work}

\subsection{Vision-based 3D Occupancy Prediction}
Vision-based 3D occupancy prediction~\cite{tian2023occ3d,tong2023scene,wang2023openoccupancy} aims to infer voxel-wise occupancy and semantics from multi-view images.
Early methods~\cite{cao2022monoscene,zhang2023occformer,huang2023tri,wei2023surroundocc,li2024bevformer} laid the foundation for learning dense 3D scene representations from images.
Subsequent studies have focused on several key challenges, including annotation costs, computational efficiency, and temporal modeling.
To reduce reliance on dense 3D annotations, RenderOcc~\cite{pan2024renderocc}, SelfOcc~\cite{huang2024selfocc}, OccNeRF~\cite{zhang2025occnerf}, GaussianOcc~\cite{gan2025gaussianocc} and DeGO~\cite{gao2026deformable} exploit neural radiance fields or 3D Gaussian representations to enable self-supervised learning.
LOcc~\cite{yu2025language} generates dense, fine-grained occupancy annotations through semantic transitive labeling, while QueryOcc~\cite{lilja2026queryocc} learns a continuous 3D occupancy field using spatiotemporal queries.
For efficient inference, FlashOcc~\cite{yu2023flashocc} and FastOcc~\cite{hou2024fastocc} replace costly 3D processing with a channel-to-height transformation, whereas VoxFormer~\cite{li2023voxformer}, SparseOcc~\cite{liu2024fully} and OPUS~\cite{wang2024opus} adopt sparse query-based representations to reduce computation.
StreamOcc~\cite{moon2025streaming}, S2GO~\cite{park2026s2go} and ViewFormer~\cite{li2024viewformer} further leverage historical observations through streaming architectures to enhance temporal feature aggregation.
More recently, OccAny~\cite{cao2026occany} and OccAnyScene~\cite{liu2026occanyscene} have explored generalizable occupancy prediction across diverse datasets. In contrast, our work focuses on injecting depth-based geometric priors into the view transformation.

\subsection{2D-to-3D View Transformation}
Transforming 2D image features into 3D representations is fundamental to 3D perception tasks~\cite{ji2025ocrfdet,ji2026depthfusion,huang2021bevdet,zhou2026generalizing}. Existing approaches can be broadly categorized into forward and backward projection paradigms.
Forward-projection methods~\cite{philion2020lift,huang2021bevdet,li2023bevdepth,li2023bevstereo,wu2025deep}, exemplified by LSS~\cite{philion2020lift}, estimate per-pixel depth distributions to lift image features along camera rays and aggregate them into BEV representations. In contrast, backward-projection methods~\cite{li2024bevformer,liu2023sparsebev,wu2026height}, such as BEVFormer~\cite{li2024bevformer}, project predefined 3D reference points onto image planes to aggregate multi-view features. To leverage the complementary strengths of both paradigms, recent studies~\cite{li2023fb,liao2025stcocc,zhu2026dr} have explored hybrid strategies. FB-Occ~\cite{li2023fb} combines forward and backward projection to construct and refine spatial representations, while COTR~\cite{ma2024cotr} integrates explicit and implicit view transformations to obtain compact 3D occupancy features. STCOcc~\cite{liao2025stcocc} further explores occupancy-guided cascading refinement to enhance spatiotemporal 3D features. More recently, Dr.Occ~\cite{zhu2026dr} leverages foundation-model depth predictions to improve the geometric alignment of voxel features. Nevertheless, conventional backward-projection methods typically rely on predefined sampling patterns that do not explicitly adapt to the non-uniform distribution of scene geometry. In contrast, our method leverages geometric priors to guide the 2D-to-3D view transformation.
% In contrast, our method back-projects predicted depth maps into 3D space to obtain geometric priors that guide the 2D-to-3D view transformation.

\section{Method}

\subsection{Overview}
\begin{figure*}[t]
    \centering
    \includegraphics[width=1\textwidth]{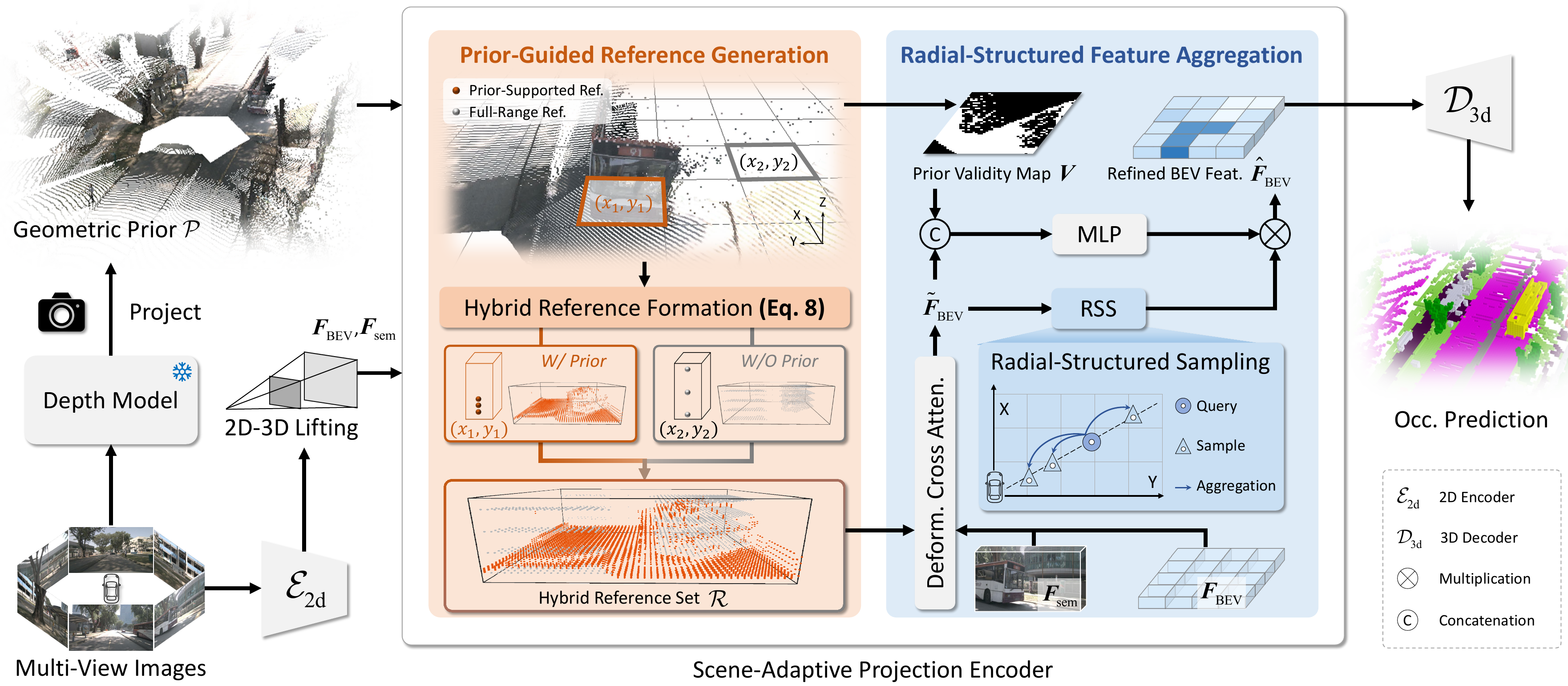}
    \caption{Overview of SAP.
    Images are encoded into semantic features and initial BEV features, while a frozen depth model produces geometric priors $\mathcal{P}$. 
    Prior-Guided Reference Generation constructs a hybrid reference set $\mathcal{R}$ from $\mathcal{P}$. 
    Radial-Structured Feature Aggregation further uses $\mathcal{R}$ to sample image features and refines BEV features through radial aggregation.
    Finally, a 3D decoder predicts occupancy.}
    \label{fig:pipeline}
\end{figure*}

As illustrated in Fig.~\ref{fig:pipeline}, a 2D encoder extracts semantic features $\boldsymbol{F}_{\mathrm{sem}}$ from images, which are subsequently lifted into BEV space to obtain initial BEV features $\boldsymbol{F}_{\mathrm{BEV}}$.
In parallel, a frozen depth model estimates metric depth maps, which are back-projected into 3D space to obtain depth-based geometric priors $\mathcal{P}$, represented as a set of 3D points.

To enable scene-adaptive view transformation, we introduce a \textbf{Scene-Adaptive Projection Encoder} comprising two complementary components. 
First, \textbf{Prior-Guided Reference Generation} constructs a hybrid reference set $\mathcal{R}$ by allocating references within prior-supported height regions for pillars with valid support and adopting full-range sampling otherwise.
Using these references, \textbf{Radial-Structured Feature Aggregation} then samples image features through deformable cross-attention, followed by radial sampling and aggregation to capture feature evidence potentially displaced by depth prediction errors.
Finally, the resulting refined BEV features $\hat{\boldsymbol{F}}_{\mathrm{BEV}}$ are fed into a 3D decoder for voxel-wise occupancy prediction.

%%%%%%%%%%%%%%%%%%%%%%%%%%%%%%%%%%%%%%%%%
%%%%%%%%%%%%%%%%%%%%%%%%%%%%%%%%%%%%%%%%%

%-------------------------------------------------------------------------
% \newpage
\subsection{BEV Encoding} 
Given multi-view images, we first extract semantic image features $\boldsymbol{F}_{\mathrm{sem}}$ using a 2D encoder~\cite{he2016deep}. Following the LSS-based paradigm~\cite{philion2020lift}, a categorical depth distribution is predicted for each image feature, based on which the features are lifted into 3D frustums and subsequently aggregated onto the BEV plane. This produces an initial BEV representation $\boldsymbol{F}_{\mathrm{BEV}}$, which serves as the BEV queries for the subsequent Scene-Adaptive Projection Encoder.

%-------------------------------------------------------------------------
\subsection{Prior-Guided Reference Generation}
\noindent\textbf{Geometric Prior Acquisition.}
We first employ a frozen depth prediction model~\cite{piccinelli2025unidepthv2} to estimate metric depth maps. Using the calibrated camera geometry, the predicted depths are back-projected into 3D and transformed into the ego coordinate system. The resulting 3D points from all views are merged to form the depth-based geometric priors $\mathcal{P}$:
\begin{equation}
\mathcal{P}
=
\{\boldsymbol{p}_n=(x_n,y_n,z_n)\}_{n=1}^{N}.
\end{equation}
We further associate each point with its corresponding BEV grid. 
Let $\mathcal{P}_i$ denote the points falling into the $i$-th grid. We then construct a BEV validity map $\boldsymbol{V}\in\{0,1\}^{H\times W}$, where $H$ and $W$ denote the spatial resolution of the discretized BEV space. Each element is defined as:
\begin{equation}
V_i
=
\mathbb{I}\left(\mathcal{P}_i\neq\varnothing\right),
\end{equation}
where $\mathbb{I}(\cdot)$ denotes the indicator function, and $V_i$ indicates whether the $i$-th BEV pillar contains points from $\mathcal{P}$.

%%%%%%%%%%%%%%%%%%%%%%%%%%%%%%%%%%%%%%%%%
\begin{figure}[t]
    \centering
    \includegraphics[width=1.0\columnwidth]{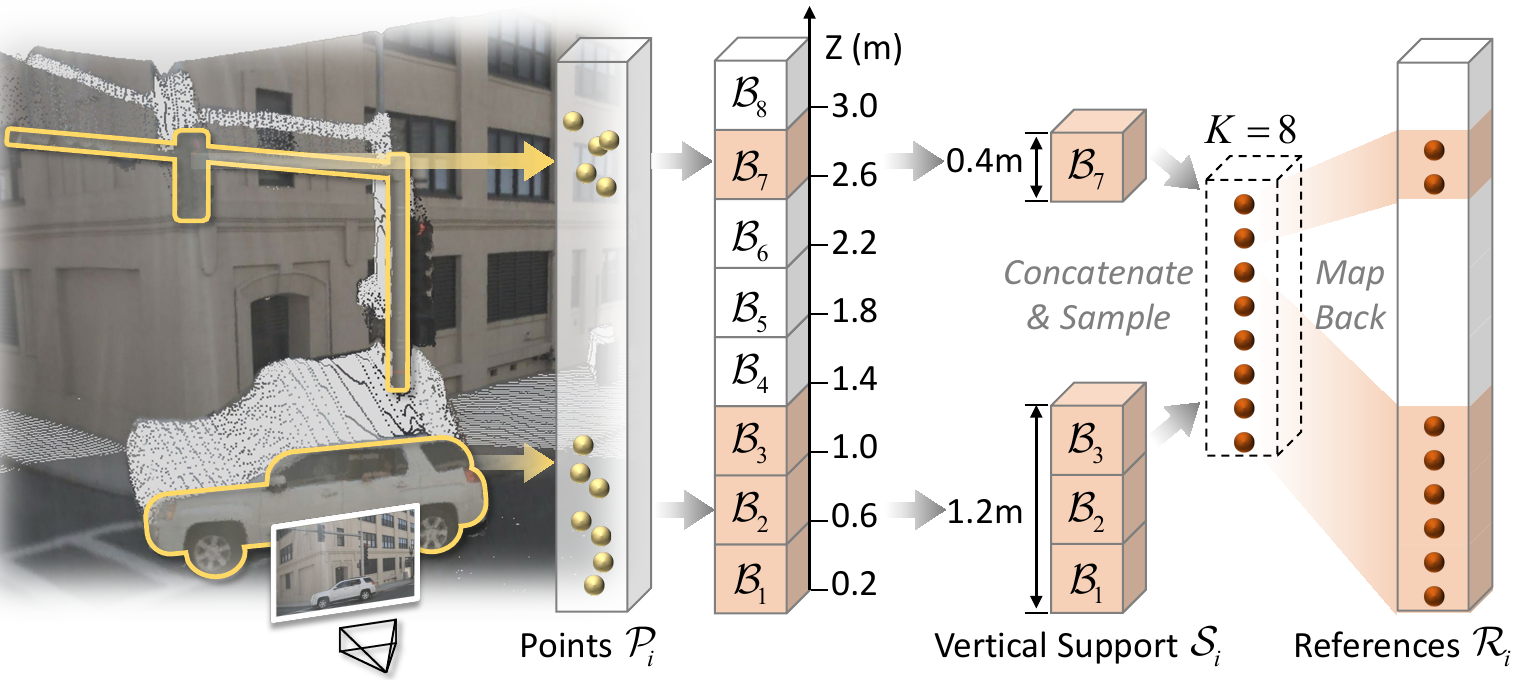}
    \caption{Illustration of prior-supported reference formation.
    Supported bins are identified from the 3D points in $\mathcal{P}$, concatenated for uniform sampling, and mapped back to their original heights.}
    \label{fig:PGRG_ref3d}
    % \vspace{-4pt}
\end{figure}
%%%%%%%%%%%%%%%%%%%%%%%%%%%%%%%%%%%%%%%%%

%%%%%%%%%%%%%%%%%%%%%%%%%%%%%%%%%%%%%%%%%
\begin{figure*}[t]
    \centering
    \includegraphics[width=1\textwidth]{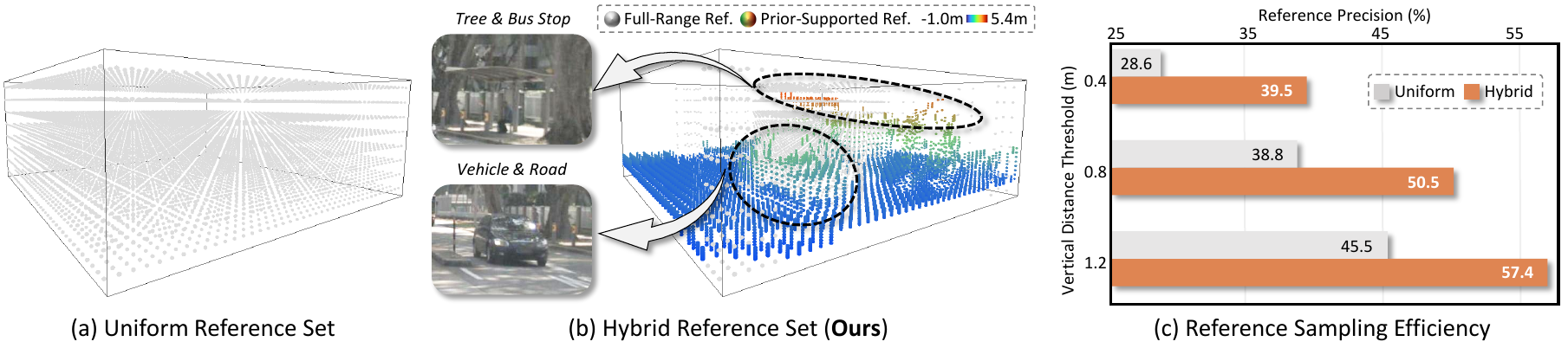}
    \caption{Comparison of uniform and hybrid reference sampling.
    (a) Uniform sampling distributes references over a predefined height range. 
    (b) Hybrid sampling allocates references within prior-supported regions, retaining full-range sampling where priors are unavailable. 
    (c) With the same sampling budget, hybrid sampling consistently achieves higher reference precision across different distance thresholds.}
    \label{fig:PGRG}
\end{figure*}
%%%%%%%%%%%%%%%%%%%%%%%%%%%%%%%%%%%%%%%%%

% 1. 根据 geometric prior 确定有几何支持的高度区间: P_i -> M_{i,b} -> S_i
% 2. 有支持区域则使用 prior-supported domain，否则使用 full range: S_i -> \Omega_i
% 3. 在选定采样域中均匀分配固定数量的 3D references: {A}_i(\Omega_i,K) -> R_i
\noindent\textbf{Hybrid Reference Formation.}
Given a fixed sampling budget of $K$ references per pillar, our goal is to allocate reference points to geometrically relevant height regions. 
To this end, we adopt a hybrid strategy that determines the vertical sampling domain based on the availability of geometric priors. 
Specifically, references are allocated within prior-supported regions when valid geometric support is available, while full-range sampling is retained otherwise.

To identify valid vertical support within the $i$-th BEV pillar, we discretize the height range $[z_{\min},z_{\max}]$ into $B$ bins of size $\Delta z$, where the $b$-th bin is defined as:
\begin{equation}
\mathcal{B}_b
=
[z_{\min}+(b-1)\Delta z,\,
z_{\min}+b\Delta z),
\quad b=1,\ldots,B.
\end{equation}
We then construct a binary indicator to identify height bins containing sufficient geometric evidence:
\begin{equation}
M_{i,b}
=
\mathbb{I}
\left(
\sum_{\boldsymbol{p}_n\in\mathcal{P}_i}
\mathbb{I}(z_n\in\mathcal{B}_b)
\geq N_{\min}
\right),
\end{equation}
where $N_{\min}$ is the minimum number of points required to activate a height bin.
The activated bin indices and corresponding vertical support are defined as:
\begin{equation}
\mathcal{I}_i
=
\{b\mid M_{i,b}=1\},
\quad
\mathcal{S}_i
=
\bigcup_{b\in\mathcal{I}_i}\mathcal{B}_b.
\end{equation}
By independently identifying supported height bins, the resulting vertical support $\mathcal{S}_i$ preserves potentially discontinuous and multi-layer vertical structures without merging separated regions into a single continuous interval.

Based on the identified vertical support, we define a unified sampling domain for each BEV pillar:
\begin{equation}
\Omega_i
=
\begin{cases}
\mathcal{S}_i, & \mathcal{S}_i\neq\varnothing,\\
[z_{\min},z_{\max}], & \mathcal{S}_i=\varnothing.
\end{cases}
\end{equation}
Given the selected sampling domain $\Omega_i$, we define a reference allocation operator $\mathcal{A}_i(\Omega_i,K)$ that uniformly distributes $K$ reference points within this domain. 
For prior-supported pillars, as illustrated in Fig.~\ref{fig:PGRG_ref3d}, the activated height bins are conceptually concatenated in ascending height order, enabling uniform allocation across discontinuous regions.
For pillars without valid support, sampling is performed over the full height range.

Let $L_i=|\Omega_i|$ denote the total length of the sampling domain, where $|\Omega_i|$ represents its one-dimensional measure.
The $K$ sampling positions are uniformly allocated as:
\begin{equation}
s_{i,k}
=
\frac{k-\frac{1}{2}}{K}L_i,
\qquad k=1,\ldots,K.
\end{equation}
Each $s_{i,k}$ is mapped back to its corresponding height $z_{i,k}$ in the original vertical domain.
Combined with the BEV pillar center $(x_i,y_i)$, this yields the 3D reference point $\boldsymbol{r}_{i,k}=(x_i,y_i,z_{i,k})$.
The final hybrid reference set $\mathcal{R}$ is obtained by combining references from all BEV pillars:
\begin{equation}
\mathcal{R} = \bigcup_i \mathcal{A}_i(\Omega_i,K).
\end{equation}

Fig.~\ref{fig:PGRG}(a) and (b) illustrate the difference between uniform and hybrid reference formation under the same sampling budget ($K=8$).
By leveraging geometric priors, our hybrid method concentrates references within geometrically supported regions while retaining full-range sampling where valid priors are unavailable. 
To quantify this advantage, we measure reference precision as the percentage of reference points whose vertical distance to the nearest occupied voxel in the same BEV pillar falls below a given threshold.
As shown in Fig.~\ref{fig:PGRG}(c), our hybrid strategy consistently achieves higher precision across different thresholds, indicating that geometric priors enable more effective reference allocation under the same sampling budget.

%%%%%%%%%%%%%%%%%%%%%%%%%%%%%%%%%%%%%%%%%
\begin{table*}[t]
\centering
\caption{Performance on Occ3D without camera visible mask during training. R denotes ResNet, and Swin denotes Swin Transformer.}
\vspace{-8pt}
\label{tab:occ3d_wo_mask}
\setlength{\tabcolsep}{8pt}
\begin{tabular}{l|cc|c|c|ccc}
\toprule
Method & Input Size & Backbone & mIoU & RayIoU
% & \multicolumn{3}{c}{RayIoU\textsubscript{1m, 2m, 4m}} \\
& RayIoU\textsubscript{1m} & RayIoU\textsubscript{2m} & RayIoU\textsubscript{4m} \\
\midrule
BEVFormer~\cite{li2024bevformer}
& $900\times1600$ & R101 & 23.7  & 32.4  & 26.1  & 32.9  & 38.0  \\
RenderOcc~\cite{pan2024renderocc}
& $512\times1408$ & Swin-B & 24.4  & 19.5  & 13.4  & 19.6  & 25.5 \\
BEVDetOcc~\cite{huang2021bevdet}
& $256\times704$ & R50 & 24.3  & 35.2  & 31.2  & 35.9  & 38.4 \\
FB-Occ~\cite{li2023fb}
& $256\times704$ & R50 & 31.1  & 39.0  & 33.0  & 40.0  & 44.0  \\
SparseOcc~\cite{liu2024fully}
& $256\times704$ & R50 & 30.9  & 36.1  & 30.2  & 36.8  & 41.2 \\
Panoptic-FlashOcc~\cite{yu2024panoptic}
& $256\times704$ & R50 & 31.6  & 38.5  & 32.8  & 39.3  & 43.4 \\
OPUS~\cite{wang2024opus}
& $256\times704$ & R50 & 36.2  & 41.2  & 34.7  & 42.1  & 46.7 \\
ALOcc-mini~\cite{chen2025alocc}
& $256\times704$ & R50 & 33.4  & 39.3  & 32.9  & 40.1  & 44.8 \\
ALOcc~\cite{chen2025alocc}
& $256\times704$ & R50 & \second{38.0}  & \second{43.7}  & \second{37.8}  & \second{44.7}  & \second{48.8}  \\
\midrule
SAP-mini
& $256\times704$ & R50 & 35.5  & 41.6  & 35.3  & 42.5  & 46.9 \\
SAP
& $256\times704$ & R50 & \first{38.9}  & \first{44.5}  & \first{38.4}  & \first{45.5}  & \first{49.5} \\

\bottomrule
\end{tabular}

% \vspace{-4pt}
\end{table*}
%%%%%%%%%%%%%%%%%%%%%%%%%%%%%%%%%%%%%%%%%

%%%%%%%%%%%%%%%%%%%%%%%%%%%%%%%%%%%%%%%%%
\begin{figure*}[t!]
    \centering
    \includegraphics[width=1\textwidth]{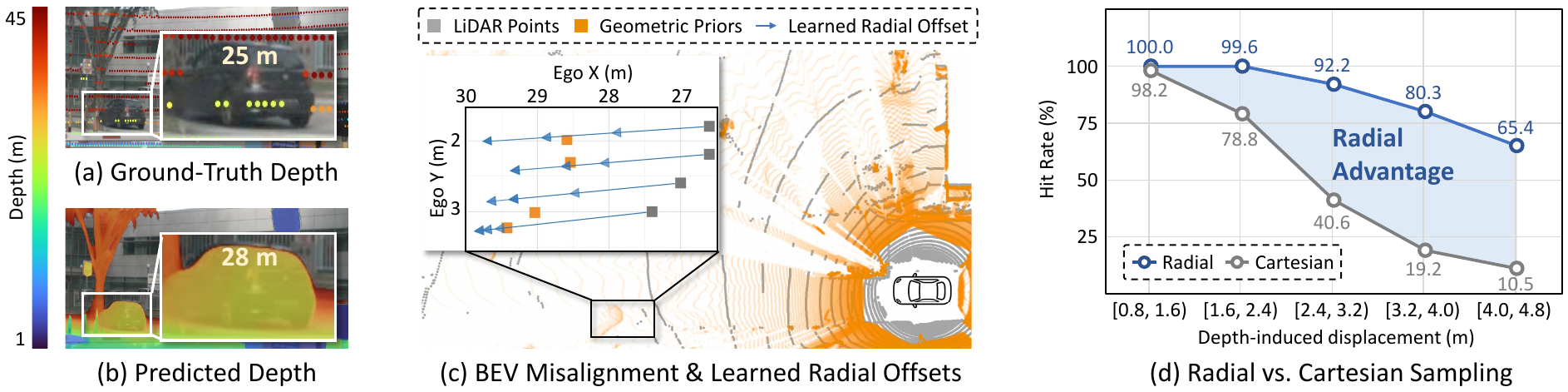}
    \caption{Analysis of radial-structured sampling.
    (a--b) Comparison of ground-truth and predicted depth for the highlighted vehicle, illustrating depth overestimation.
    (c) Corresponding BEV visualization showing depth-induced misalignment and learned radial offsets directed toward displaced points in $\mathcal{P}$.
    (d) Comparison of radial and Cartesian sampling across different displacement ranges. A hit is counted when a sampling point falls within 0.8\,m of the target BEV location. Radial offsets consistently achieve higher hit rates.}    
    % \vspace{-5pt}
    \label{fig:RSFA_car_misalign}
\end{figure*}
%%%%%%%%%%%%%%%%%%%%%%%%%%%%%%%%%%%%%%%%%

% Depth Prediction Error → BEV Spatial Misalignment → Approximate Radial Structure → Radial Sampling
%-------------------------------------------------------------------------
\subsection{Radial-Structured Feature Aggregation}
Given the hybrid reference set $\mathcal{R}$, we employ deformable cross-attention~\cite{li2024bevformer} to update the initial BEV representation $\boldsymbol{F}_{\mathrm{BEV}}$. Specifically, the 3D references are projected onto the image planes using calibrated camera parameters to guide feature sampling. The sampled image features are then aggregated to update the BEV queries, yielding:
\begin{equation}
\tilde{\boldsymbol{F}}_{\mathrm{BEV}}
=
\phi
\left(
\boldsymbol{F}_{\mathrm{BEV}},
\boldsymbol{F}_{\mathrm{sem}},
\mathcal{R}
\right),
\end{equation}
where $\phi(\cdot)$ denotes the deformable cross-attention operator. 

\noindent\textbf{Radial-Structured Sampling.}
Although geometric priors provide valuable guidance for reference formation, depth prediction errors can still introduce spatial misalignment in BEV features. Specifically, depth errors displace back-projected 3D points along their corresponding camera rays. 
As illustrated in Fig.~\ref{fig:RSFA_car_misalign}(a--c), depth overestimation causes the corresponding points in $\mathcal{P}$ to deviate from the actual positions of the highlighted vehicle. 
Since surround-view cameras are mounted close to the ego center, such displacement approximately follows the ego-centered radial direction in BEV space.
Motivated by this observation, we introduce radial-structured sampling to explore multiple candidate locations along this direction, enabling BEV queries to aggregate potentially misaligned feature evidence.

For the $i$-th BEV grid centered at $(x_i,y_i)$, we define its radial distance and azimuth as:
\begin{equation}
    d_{i}
    =
    \sqrt{x_i^2+y_i^2},
    \qquad
    \theta_{i}
    =
    \operatorname{atan2}(y_i,x_i).
\end{equation}
Rather than predicting unconstrained Cartesian offsets, we predict $J$ query-dependent signed radial offsets $\{\Delta d_{i,j}\}_{j=1}^{J}$ from $\tilde{\boldsymbol{F}}_{\mathrm{BEV}}$. The $j$-th sampling location is defined as:
\begin{equation}
\boldsymbol{\xi}_{i,j}
=
\left(
(d_{i}+\Delta d_{i,j})\cos\theta_{i},\,
(d_{i}+\Delta d_{i,j})\sin\theta_{i}
\right).
\end{equation}
Positive and negative offsets allow each query to explore locations farther from or closer to the ego center, respectively. 
Meanwhile, attention weights $\{\alpha_{i,j}\}_{j=1}^{J}$ are predicted from the corresponding BEV query and normalized across the sampling locations.
The resulting radial feature update is:
\begin{equation}
\Delta\boldsymbol{F}_{\mathrm{rad}}(i)
=
\sum_{j=1}^{J}
\alpha_{i,j}\,
\psi
\left(
\tilde{\boldsymbol{F}}_{\mathrm{BEV}},
\boldsymbol{\xi}_{i,j}
\right),
\end{equation}
where $\psi(\cdot)$ denotes feature sampling from $\tilde{\boldsymbol{F}}_{\mathrm{BEV}}$.

By aggregating features from multiple radial locations, each BEV query can incorporate feature evidence potentially displaced by depth prediction errors.
As visualized in Fig.~\ref{fig:RSFA_car_misalign}(c), the learned radial offsets extend toward points in $\mathcal{P}$, illustrating how radial sampling explores candidate locations associated with spatial misalignment.
To quantify this capability, Fig.~\ref{fig:RSFA_car_misalign}(d) further compares the hit rates of radial and Cartesian sampling, where a target is considered hit if at least one sampling point falls within 0.8\,m of its displaced BEV location.
Radial sampling consistently achieves higher hit rates, indicating its effectiveness in exploring potentially misaligned feature evidence.

%%%%%%%%%%%%%%%%%%%%%%%%%%%%%%%%%%%%%%%%%%
\begin{table*}[t]
\centering
\Huge
\caption{Performance on Occ3D with the camera visibility mask during training.}
\vspace{-8pt}
\resizebox{1.0\textwidth}{!}{
\begin{tabular}{l|cc|c|ccccccccccccccccc}

\toprule
Method
& Input Size
& Backbone
& mIoU
& \begin{sideways}{\textcolor{otherscolor}{$\blacksquare$} others}\end{sideways}  
& \begin{sideways}{\textcolor{barriercolor}{$\blacksquare$} barrier}\end{sideways}  
& \begin{sideways}{\textcolor{bicyclecolor}{$\blacksquare$} bicycle}\end{sideways}  
& \begin{sideways}{\textcolor{buscolor}{$\blacksquare$} bus}\end{sideways}  
& \begin{sideways}{\textcolor{carcolor}{$\blacksquare$} car}\end{sideways}  
& \begin{sideways}{\textcolor{constructcolor}{$\blacksquare$} cons. veh.}\end{sideways}  
& \begin{sideways}{\textcolor{motorcolor}{$\blacksquare$} motorcycle}\end{sideways}  
& \begin{sideways}{\textcolor{pedestriancolor}{$\blacksquare$} pedestrian}\end{sideways}  
& \begin{sideways}{\textcolor{trafficcolor}{$\blacksquare$} traffic cone}\end{sideways}  
& \begin{sideways}{\textcolor{trailercolor}{$\blacksquare$} trailer}\end{sideways}  
& \begin{sideways}{\textcolor{truckcolor}{$\blacksquare$} truck}\end{sideways}  
& \begin{sideways}{\textcolor{drivablecolor}{$\blacksquare$} drive. surf.}\end{sideways}  
& \begin{sideways}{\textcolor{otherflatcolor}{$\blacksquare$} other flat}\end{sideways}  
& \begin{sideways}{\textcolor{sidewalkcolor}{$\blacksquare$} sidewalk}\end{sideways}  
& \begin{sideways}{\textcolor{terraincolor}{$\blacksquare$} terrain}\end{sideways}  
& \begin{sideways}{\textcolor{manmadecolor}{$\blacksquare$} manmade}\end{sideways}  
& \begin{sideways}{\textcolor{vegetationcolor}{$\blacksquare$} vegetation}\end{sideways} 
\\ 
\midrule

BEVFormer~\cite{li2024bevformer} & $900\times1600$ & R101 & 39.2  & 5.0  & 44.9  & 26.2  & \first{59.7}  & 55.1  & 27.9  & 29.1  & \second{34.3}  & 29.6  & 29.1  & \first{50.5}  & 44.4  & 22.4  & 21.5  & 19.5  & 39.3  & 31.1   \\ 
CTF-Occ~\cite{tian2023occ3d} & $900\times1600$ & R101 & 28.5  & 8.1  & 39.3  & 20.6  & 38.3  & 42.2  & 16.9  & 24.5  & 22.7  & 21.1  & 23.0  & 31.1  & 53.3  & 33.8  & 38.0  & 33.2  & 20.8  & 18.0   \\
TPVFormer~\cite{huang2023tri} & $900\times1600$ & R101 & 27.8  & 7.2  & 38.9  & 13.7  & 40.8  & 45.9  & 17.2  & 20.0  & 18.9  & 14.3  & 26.7  & 34.2  & 55.7  & 35.5  & 37.6  & 30.7  & 19.4  & 16.8   \\ 
OSP~\cite{shi2024occupancy} & $900\times1600$ & R101 & 41.2  & 11.0  & 49.0  & 27.7  & 50.2  & \second{56.0}  & 23.0  & 31.0  & 30.9  & 30.3  & 35.6  & 41.2  & 82.1  & 42.6  & 51.9  & 55.1  & 44.8  & 38.2   \\ 
FastOcc~\cite{hou2024fastocc} & $640\times1600$ & R101 & 39.2  & 12.1  & 43.5  & 28.0  & 44.8  & 52.2  & 23.0  & 29.1  & 29.7  & 27.0  & 30.8  & 38.4  & 82.0  & 41.9  & 51.9  & 53.7  & 41.0  & 35.5   \\ 
PanoOcc~\cite{wang2024panoocc} & $640\times1600$ & R101 & 42.1  & 11.7  & 50.5  & 29.6  & 49.4  & 55.5  & 23.3  & \second{33.3}  & 30.6  & 31.0  & 34.4  & 42.6  & 83.3  & 44.2  & 54.4  & 56.0  & 45.9  & 40.4   \\
BEVDetOcc~\cite{huang2021bevdet} & $256\times704$ & Swin-B & 42.0  & 12.2  & 49.6  & 25.1  & \second{52.0}  & 54.5  & 27.9  & 28.0  & 28.9  & 27.2  & 36.4  & 42.2  & 82.3  & 43.3  & 54.6  & 57.9  & 48.6  & 43.6   \\ 
FlashOcc~\cite{yu2023flashocc}  & $256\times704$ & R50 & 32.0  & 6.2  & 39.6  & 11.3  & 36.3  & 44.0  & 16.3  & 14.7  & 16.9  & 15.8  & 28.6  & 30.9  & 78.2  & 37.5  & 47.4  & 51.4  & 36.8  & 31.4   \\ 
FB-Occ~\cite{li2023fb} & $256\times704$ & R50 & 39.8  & 13.8  & 44.5  & 27.1  & 46.2  & 49.7  & 24.6  & 27.4  & 28.5  & 28.2  & 33.7  & 36.5  & 81.7  & 44.1  & 52.6  & 56.9  & 42.6  & 38.1   \\ 
DHD~\cite{wu2025deep} & $256\times704$ & R50 & 41.5  & 12.7  & 48.7  & 26.3  & 43.2  & 52.9  & 27.3  & 28.5  & 28.5  & 30.0  & 35.8  & 40.2  & 83.1  & 44.7  & 54.7  & 57.7  & 48.9  & 42.1 \\
SA-Occ~\cite{chen2025sa} & $256\times704$ & R50 & 41.7  & 11.9  & 50.5  & 22.1  & 45.0  & 54.2  & 25.9  & 24.0  & 24.4  & 26.3  & 35.4  & 41.0  & 84.3  & 46.1  & 57.0  & 61.2  & 52.8  & 46.8 \\
COTR~\cite{ma2024cotr}  & $256\times704$ & R50 & 44.5  & 13.3  & 52.1  & 32.0  & 46.0  & 55.6  & \second{32.6}  & 32.8  & 30.4  & 34.1  & 37.7  & 41.8  & 84.5  & 46.2  & 57.6  & 60.7  & 52.0  & 46.3   \\ 
STCOcc~\cite{liao2025stcocc}  & $256\times704$ & R50 &  44.6  & 15.3  & \second{52.9}  & 31.6  & 46.4  & 55.9  & 31.5  & 32.6  & 32.1  & 34.5  & \second{39.5}  & 42.5  & 83.6  & 47.8  & 56.4  & 60.1  & 50.8  & 44.8   \\ 
Dr.Occ~\cite{zhu2026dr}& $256\times704$ & R50 & 44.1  & 14.0  & \first{53.0}  & \second{32.6}  & 45.8  & 54.9  & 29.7  & 31.5  & 30.9  & 35.7  & 38.0  & 42.5  & 83.8  & 44.9  & 56.1  & 58.8  & 51.1  & 45.6   \\
FDROcc~\cite{chen2026fdr} & $256\times704$ & R50 & 41.2  & 13.6 & 46.5 & 28.3 & 43.4 & 53.0 & 26.8 & 29.1 & 29.6 & 31.6 & 36.4 & 39.9 & 81.6 & 44.3 & 53.0 & 55.5 & 47.2 & 40.6 \\
ALOcc-mini~\cite{chen2025alocc} & $256\times704$ & R50 & 41.4  & 14.2  & 48.6  & 28.7  & 44.8  & 52.8  & 24.7  & 29.2  & 29.0  & 32.0  & 34.6  & 39.6  & 82.4  & 46.9  & 54.8  & 57.7  & 44.7  & 39.3   \\ 
ALOcc~\cite{chen2025alocc} & $256\times704$ & R50 & \second{45.5}  & \second{15.3}  & 52.5  & 30.8  & 47.2  & 55.9  & \first{32.7}  & \second{33.3}  & 32.4  & \second{36.2}  & 38.9  & 43.7  & \first{84.9}  & \second{48.5}  & \first{58.8}  & \first{61.9}  & \second{53.5}  & \second{47.3}   \\ 
\hline
SAP-mini & $256\times704$ & R50 & 43.5  & 14.6  & 51.3  & 31.0  & 47.4  & \second{56.0}  & 26.7  & \second{33.3}  & 33.0  & 34.5  & 36.9  & 42.9  & 82.8  & 47.4  & 54.9  & 58.0  & 47.5  & 41.8 \\
SAP & $256\times704$ & R50 & \first{46.6}  & \first{15.7}  & 52.7  & \first{33.1}  & 49.3  & \first{57.7}  & 32.0  & \first{34.9}  & \first{34.6}  & \first{38.3}  & \first{40.9}  & \second{47.2}  & \second{84.7}  & \first{49.4}  & \second{57.8}  & \second{61.6}  & \first{54.6}  & \first{47.7} \\ 

\bottomrule
\end{tabular}}
\label{tab:Occ3d_w_camera_mask}
% \vspace{-3pt}
\end{table*}
%%%%%%%%%%%%%%%%%%%%%%%%%%%%%%%%%%%%%%%%%

\noindent\textbf{Prior-Conditioned Gating.}
Given the BEV validity map $\boldsymbol{V}$ and projected BEV feature $\tilde{\boldsymbol{F}}_{\mathrm{BEV}}$, we predict a spatial gate $\boldsymbol{G}\in[0,1]^{H\times W}$ using a lightweight MLP $\eta$:
\begin{equation}
\boldsymbol{G} = \sigma \left( \eta \left( \tilde{\boldsymbol{F}}_{\mathrm{BEV}} \mathbin{\Vert} \boldsymbol{V} \right) \right),
\end{equation}
where $\mathbin{\Vert}$ denotes channel-wise concatenation and $\sigma(\cdot)$ is the sigmoid function. The radial-aggregated feature is then adaptively injected through a residual update:
\begin{equation}
\hat{\boldsymbol{F}}_{\mathrm{BEV}}
=
\tilde{\boldsymbol{F}}_{\mathrm{BEV}}
+
\boldsymbol{G}
\odot
\Delta\boldsymbol{F}_{\mathrm{rad}},
\end{equation}
where $\odot$ denotes element-wise multiplication. We refer to
$\hat{\boldsymbol{F}}_{\mathrm{BEV}}$ as the refined BEV feature, which is subsequently decoded for 3D occupancy prediction.

%-------------------------------------------------------------------------
\subsection{Loss Function}
Following~\cite{chen2025alocc,liu2024fully}, we supervise the query-based 3D decoder using classification, mask, and dice objectives. The total training loss is formulated as:
\begin{equation}
\mathcal{L}
=
\lambda_{\mathrm{cls}}\mathcal{L}_{\mathrm{cls}}
+
\lambda_{\mathrm{mask}}\mathcal{L}_{\mathrm{mask}}
+
\lambda_{\mathrm{dice}}\mathcal{L}_{\mathrm{dice}},
\end{equation}
where $\mathcal{L}_{\mathrm{cls}}$ is the cross-entropy loss for query classification, $\mathcal{L}_{\mathrm{mask}}$ is the cross-entropy loss for occupancy mask prediction, and $\mathcal{L}_{\mathrm{dice}}$ measures the overlap between predicted and ground-truth masks. We set the corresponding loss weights to $\lambda_{\mathrm{cls}}=2.0$, $\lambda_{\mathrm{mask}}=5.0$, and $\lambda_{\mathrm{dice}}=5.0$.

\section{Experiments}

%%%%%%%%%%%%%%%%%%%%%%%%%%%%%%%%%%%%%%%%%
\begin{figure*}[t!]
    \centering
    \includegraphics[width=1\textwidth]{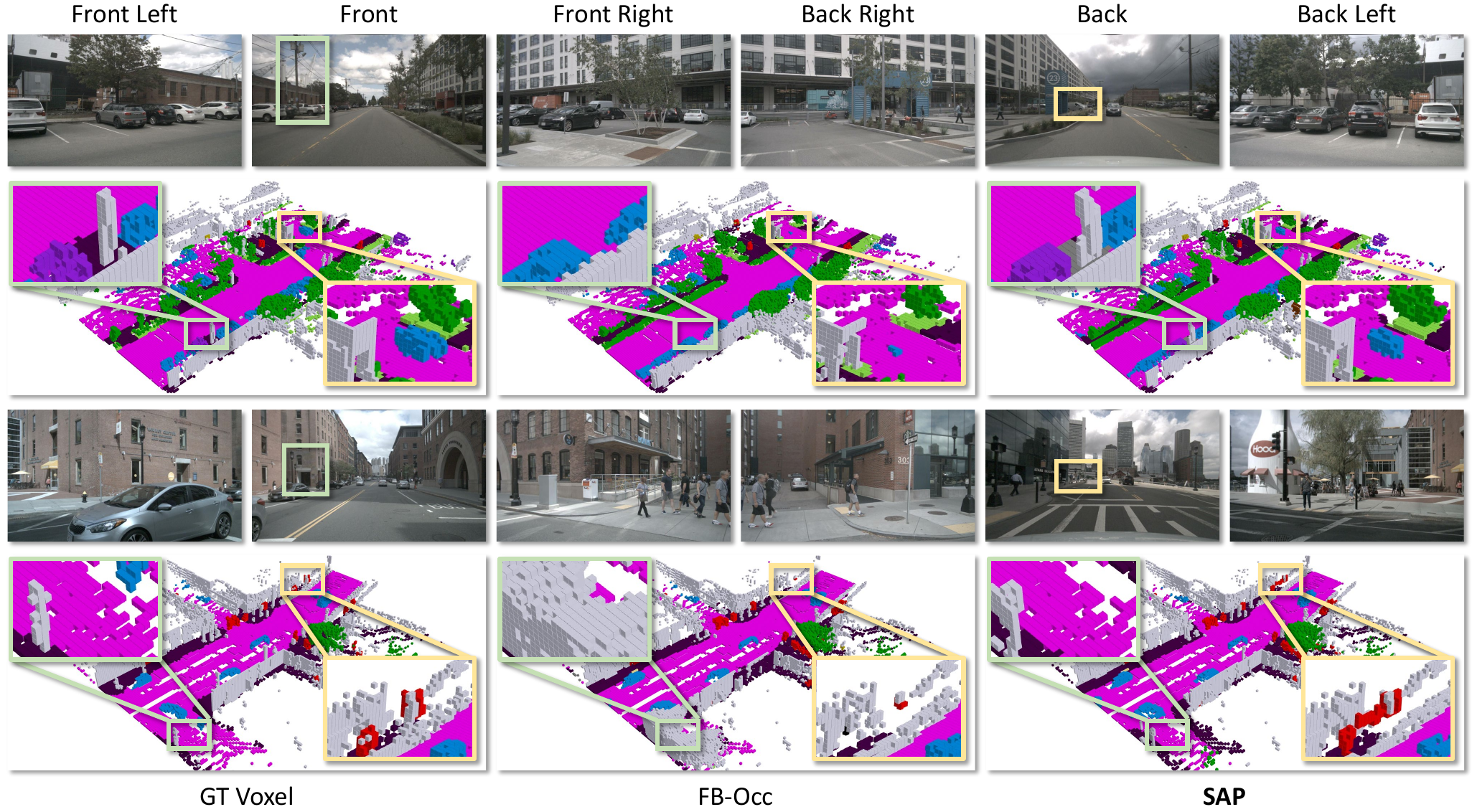}
    \vspace{-10pt}
    \caption{Visual comparisons of FB-Occ and our SAP on the validation set of the Occ3D dataset.}
    \label{fig:occ_vis}
\end{figure*}
%%%%%%%%%%%%%%%%%%%%%%%%%%%%%%%%%%%%%%%%%

%%%%%%%%%%%%%%%%%%%%%%%%%%%%%%%%%%%%%%%%%
\begin{figure*}[t]
\centering

% ===== Left: Table =====
\begin{minipage}[t]{0.49\textwidth}
    \vspace{0pt}
    \centering
    \captionof{table}{Ablation of SAP components.
    PGRG: Prior-Guided Reference Generation;
    RSFA: Radial-Structured Feature Aggregation.}
    \label{tab:component_ablation}
    \vspace{-5pt}

    \setlength{\tabcolsep}{12pt}
    \begin{tabular}{lccc}
        \toprule
        Model & PGRG & RSFA & mIoU \\
        \midrule
        Baseline & & & 41.43 \\
        i   & \checkmark & & 42.78 (+1.35) \\
        ii  & & \checkmark & 42.48 (+1.05) \\
        iii & \checkmark & \checkmark &
        \textbf{43.53 (+2.10)} \\
        \bottomrule
    \end{tabular}
\end{minipage}
\hfill
% ===== Right: Figure =====
\begin{minipage}[t]{0.49\textwidth}
    \vspace{0pt}
    \centering
    \includegraphics[width=\linewidth]{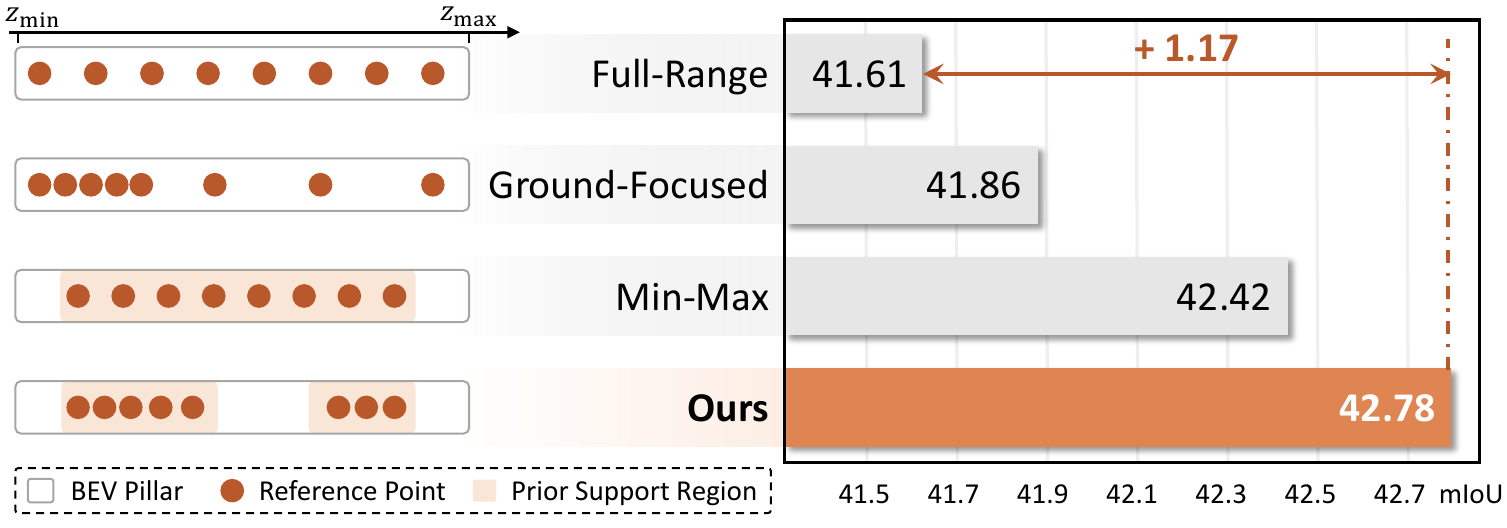}
    \captionof{figure}{Ablation of different reference formations in PGRG.}
    \label{fig:ablation_PGRG_ref_formation}
\end{minipage}
\vspace{-5pt}
\end{figure*}
%%%%%%%%%%%%%%%%%%%%%%%%%%%%%%%%%%%%%%%%%

\subsection{Experimental Setup}
\noindent\textbf{Datasets and Metrics.} 
We evaluate our method on the nuScenes dataset~\cite{caesar2020nuscenes}, which consists of 700 training scenes and 150 validation scenes annotated at 2~Hz. Based on nuScenes, Occ3D~\cite{tian2023occ3d} provides voxel-level occupancy annotations within a spatial range of $[-40,40]$~m along the $X$ and $Y$ axes and $[-1,5.4]$~m along the $Z$ axis, with a voxel size of $0.4$~m. 
The benchmark defines 18 semantic classes, where occupied voxels are categorized into 17 semantic classes, including 16 common categories and an additional ``others'' class. Following prior works~\cite{chen2025alocc,liao2025stcocc}, we report mean Intersection over Union (mIoU) for models trained with the camera visibility mask, and both mIoU and RayIoU for those trained without the mask.
% Following prior work~\cite{chen2025alocc}, we report mean Intersection over Union (mIoU) for models trained with the camera visibility mask, and evaluate RayIoU for models trained without this mask. 
% RayIoU evaluates occupancy along rays at three distance thresholds (1m, 2m, 4m), with their average as the final score.
% RayIoU evaluates occupancy along rays at three distance thresholds (1m, 2m, 4m), with the mean of these results serving as the final score.

\noindent\textbf{Implementation Details.}
We develop two variants, SAP-mini and SAP, both based on the ALOcc~\cite{chen2025alocc} series. SAP-mini adopts a reduced channel width for a lightweight configuration. UniDepth-V2~\cite{piccinelli2025unidepthv2} is utilized as the frozen depth prediction model for both variants. All models are optimized with AdamW~\cite{loshchilov2017decoupled} using an initial learning rate of $2\times10^{-4}$ and a total batch size of 16. Our implementation is built upon MMDetection3D~\cite{chen2019mmdetection}, and all experiments are conducted on four NVIDIA GeForce RTX 4090 GPUs.

\subsection{Comparison with the State-of-the-Art}

\noindent\textbf{Occ3D without Camera Visible Mask.} As shown in Tab.~\ref{tab:occ3d_wo_mask}, SAP achieves the best performance, with 38.9 mIoU and 44.5 RayIoU. Compared with the second-best ALOcc~\cite{chen2025alocc}, SAP improves mIoU and RayIoU by 0.9 and 0.8 points, respectively. SAP outperforms OPUS~\cite{wang2024opus} by 2.7 points in mIoU and 3.3 points in RayIoU, while maintaining consistent advantages across all RayIoU thresholds.

\noindent\textbf{Occ3D with Camera Visible Mask.}
Tab.~\ref{tab:Occ3d_w_camera_mask} shows that SAP achieves state-of-the-art performance, surpassing the second-best method ALOcc~\cite{chen2025alocc} by 1.1 points and FB-Occ~\cite{li2023fb} by 6.8 points. Fig.~\ref{fig:occ_vis} further illustrates these improvements. Specifically, while FB-Occ misclassifies a car as a truck and produces incomplete pedestrian structures, SAP achieves more accurate semantics and preserves more complete geometry. Further analysis of computational cost is provided in the supplementary material.

%%%%%%%%%%%%%%%%%%%%%%%%%%%%%%%%%%%%%%%%%
\begin{figure*}[t]
\centering
% ===== Left figure =====
\begin{minipage}[t]{0.48\textwidth}
    \centering
    \includegraphics[width=\linewidth]{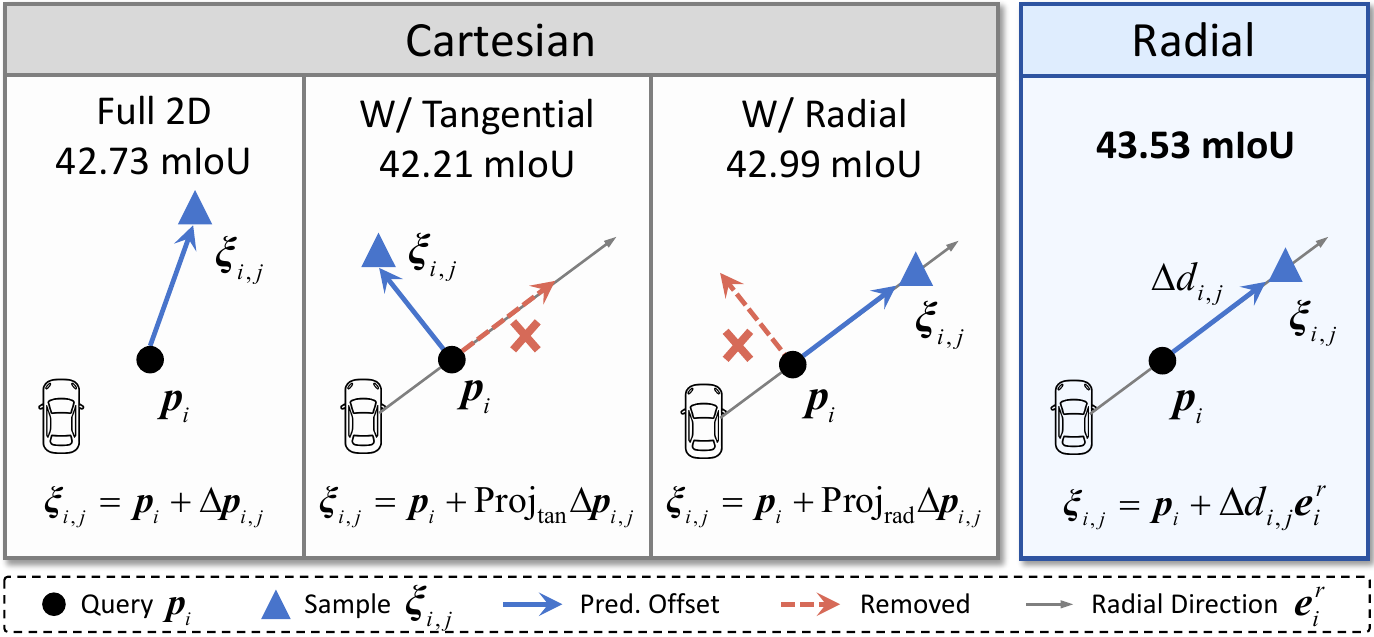}
    \caption{Ablation of different sampling offsets in RSFA.}
    \label{fig:ablations_RSFA_offsets}
\end{minipage}
\hfill
% ===== Right figure =====
\begin{minipage}[t]{0.48\textwidth}
    \centering
    \includegraphics[width=\linewidth]{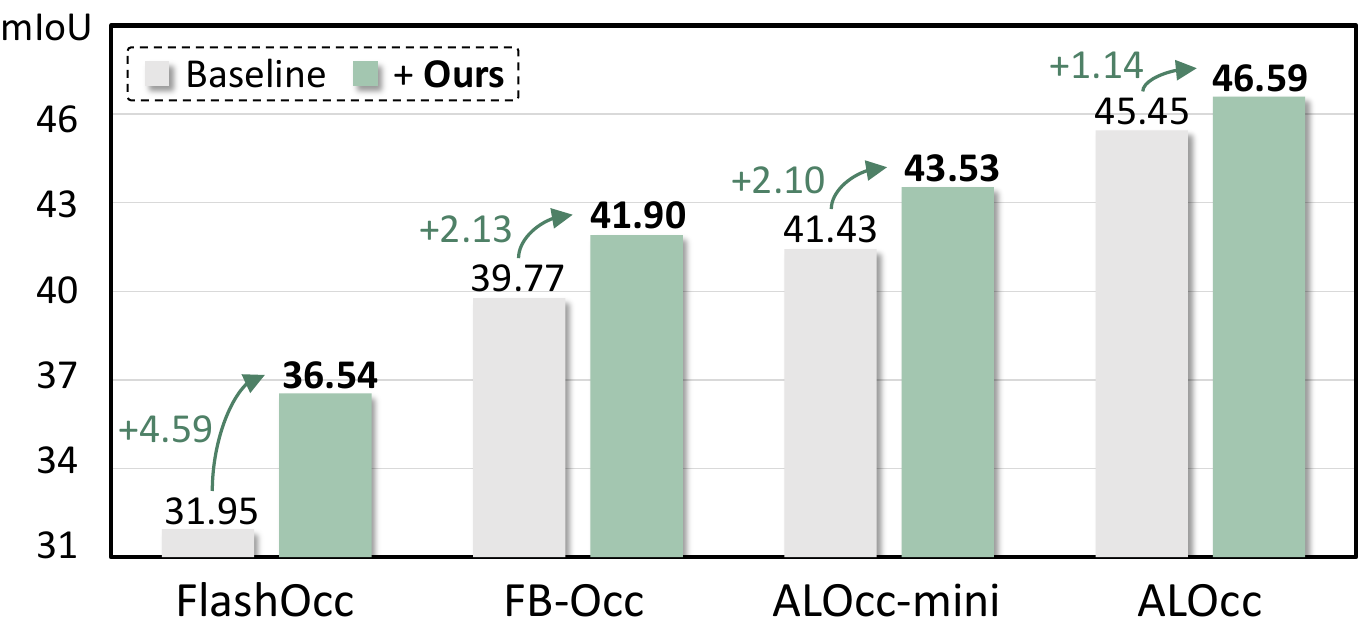}
    \caption{Generalization on different occupancy models.}
    \label{fig:generalization}
\end{minipage}
% \vspace{-3pt}
\end{figure*}
%%%%%%%%%%%%%%%%%%%%%%%%%%%%%%%%%%%%%%%%%

%%%%%%%%%%%%%%%%%%%%%%%%%%%%%%%%%%%%%%%%%
\begin{table*}[t]
\centering
\renewcommand{\arraystretch}{1.1}

% Left: Depth Model Ablation
\begin{minipage}[t]{0.48\textwidth}
    \vspace{0pt}
    \centering
    \caption{Ablation of different depth prediction models in PGRG.}
    \label{tab:depth_model_ablation}
    \vspace{-8pt}
    \setlength{\tabcolsep}{9pt}
    \begin{tabular}{lccc}
        \toprule
        Model & DA-V3 & MoGe-2 & UniDepth-V2 \\
        \midrule
        MAE (m)  & 2.43 & 2.02  & \textbf{1.53} \\
        RMSE (m) & 4.68 & 4.11  & \textbf{3.79} \\
        $\delta_1$ (\%)  & 76.08 & 81.86  & \textbf{88.66} \\ \hline
        mIoU     & 41.89 & 42.46  & \textbf{42.78} \\
        \bottomrule
    \end{tabular}
\end{minipage}
\hfill
% Right: Gating Ablation
\begin{minipage}[t]{0.48\textwidth}
    \vspace{0pt}
    \centering
    \caption{Ablation of gating inputs in RSFA.}
    \label{tab:rsfa_gate_ablation}
    \vspace{-8pt}
    \setlength{\tabcolsep}{12pt}
    \begin{tabular}{ccc}
        \toprule
        Query Feature & Validity Map & mIoU \\
        \midrule
         &  & 43.24 \\
        \checkmark &  & 43.13 \\
         & \checkmark & 43.20 \\
        \checkmark & \checkmark & \textbf{43.53} \\
        \bottomrule
    \end{tabular}
\end{minipage}

\end{table*}
%%%%%%%%%%%%%%%%%%%%%%%%%%%%%%%%%%%%%%%%%

\subsection{Ablation Study}
All experiments reported in this section are trained and evaluated on the Occ3D dataset with the camera visible mask.

\noindent\textbf{SAP Components.} Tab.~\ref{tab:component_ablation} presents the ablation study of the Prior-Guided Reference Generation (PGRG) and Radial-Structured Feature Aggregation (RSFA), with ALOcc-mini serving as the baseline. Starting from the baseline mIoU of 41.43, introducing PGRG improves the performance to 42.78, yielding a gain of 1.35 points. 
When applied independently, RSFA improves the mIoU to 42.48, bringing a gain of 1.05 points, indicating that radial-structured feature aggregation is beneficial even without PGRG. Finally, jointly employing PGRG and RSFA further improves the performance to 43.53 mIoU, corresponding to an overall gain of 2.10 points over the baseline model. 

\noindent\textbf{PGRG.}
We investigate PGRG from two perspectives: reference formation
strategies and depth prediction models.

First, we compare different reference formation strategies, as illustrated in Fig.~\ref{fig:ablation_PGRG_ref_formation}. Starting from the ALOcc-mini baseline (41.43 mIoU), Full-Range uniformly distributes  references over the predefined height range, while Ground-Focused allocates half of the references to the near-ground region ([0, 1] m). 
The improvement of Ground-Focused over Full-Range indicates the benefit of prioritizing geometrically relevant regions, although its fixed sampling pattern cannot adapt to varying scene geometry. 
By introducing depth-based geometric priors, Min-Max further improves mIoU to 42.42. However, it assumes continuous geometric support between the minimum and maximum heights of geometric priors, potentially assigning references to unsupported regions. In contrast, our strategy identifies potentially discontinuous vertical support regions and allocates references within them, retaining full-range sampling where valid support is unavailable. This strategy better captures the vertical distribution of scene geometry, achieving a 1.17 mIoU improvement over Full-Range.

Second, we evaluate the influence of depth models~\cite{lin2025depth,wang2026moge,piccinelli2025unidepthv2}, as reported in Tab.~\ref{tab:depth_model_ablation}. Following the configuration in Tab.~\ref{tab:component_ablation}(i), we vary only the depth model. Among the evaluated models, UniDepth-V2~\cite{piccinelli2025unidepthv2} achieves the lowest depth errors (1.53 m MAE and 3.79 m RMSE) and the highest mIoU of 42.78. These results suggest that more accurate depth predictions provide more reliable geometric support, enabling PGRG to better allocate references accordingly.

\noindent\textbf{RSFA.}
We analyze the effectiveness of radial sampling by comparing different offset parameterizations, as shown in Fig.~\ref{fig:ablations_RSFA_offsets}. Starting from Cartesian offsets (42.73 mIoU), removing the radial component decreases performance to 42.21, while removing the tangential component improves it to 42.99. Directly predicting radial offsets further achieves 43.53 mIoU, demonstrating the effectiveness in aggregating spatially misaligned features.

We further investigate the gate design by ablating its inputs, including the BEV query and validity map. Without gating, the model achieves 43.24 mIoU, while query-only and validity-only gating yield 43.13 and 43.20, respectively. Combining both inputs achieves the highest mIoU of 43.53, as the BEV query provides feature context while the validity map indicates the availability of geometric support, enabling more effective modulation of feature aggregation.
\section{Generalization}
To evaluate the generalization of SAP, we integrate it into FlashOcc~\cite{yu2023flashocc}, FB-Occ~\cite{li2023fb}, and the ALOcc series~\cite{chen2025alocc}. 
As shown in Fig.~\ref{fig:generalization}, SAP consistently improves performance across these frameworks, achieving mIoU gains of 4.59 and 2.13 points on FlashOcc and FB-Occ, respectively. It also improves ALOcc-mini from 41.43 to 43.53 mIoU and ALOcc from 45.45 to 46.59 mIoU. These results demonstrate the generalizability of SAP across different model architectures and model capacities.
\section{Conclusion}

In this paper, we present SAP, a scene-adaptive projection framework for 3D occupancy prediction. Specifically, we back-project depth maps from a frozen depth model into 3D to obtain geometric priors, which are incorporated through a Scene-Adaptive Projection Encoder. Within this encoder, we design Prior-Guided Reference Generation to construct a hybrid reference set that directs more references toward geometrically supported regions rather than empty space. Furthermore, to account for potential misalignment caused by depth estimation errors, we introduce Radial-Structured Feature Aggregation to explore multiple radial candidate locations and aggregate potentially displaced BEV features. Extensive experiments demonstrate the superior performance and strong generalizability of SAP.

\section{LLM Usage Statement}

OpenAI ChatGPT was used for language editing and grammatical refinement throughout the manuscript. All AI-assisted text was reviewed and verified by the authors, who take full responsibility for the final content.

{
    \small
    \bibliographystyle{ieeenat_fullname}
    \bibliography{main}

@String(CVPR= {IEEE Conf. Comput. Vis. Pattern Recog.})

@String(ICCV= {Int. Conf. Comput. Vis.})

@String(AAAI = {AAAI})

@String(CVPR  = {CVPR})

@String(ICCV  = {ICCV})

@inproceedings{cao2022monoscene,
  title={Monoscene: Monocular 3d semantic scene completion},
  author={Cao, Anh-Quan and De Charette, Raoul},
  booktitle={2022 IEEE/CVF Conference on Computer Vision and Pattern Recognition (CVPR)},
  pages={3981--3991},
  year={2022},
  organization={IEEE}
}

@inproceedings{zhang2023occformer,
  title={Occformer: Dual-path transformer for vision-based 3d semantic occupancy prediction},
  author={Zhang, Yunpeng and Zhu, Zheng and Du, Dalong},
  booktitle={2023 IEEE/CVF International Conference on Computer Vision (ICCV)},
  pages={9399--9409},
  year={2023},
  organization={IEEE}
}

@inproceedings{huang2023tri,
  title={Tri-perspective view for vision-based 3d semantic occupancy prediction},
  author={Huang, Yuanhui and Zheng, Wenzhao and Zhang, Yunpeng and Zhou, Jie and Lu, Jiwen},
  booktitle={2023 IEEE/CVF Conference on Computer Vision and Pattern Recognition (CVPR)},
  pages={9223--9232},
  year={2023},
  organization={IEEE}
}

@inproceedings{gan2025gaussianocc,
  title={Gaussianocc: Fully self-supervised and efficient 3d occupancy estimation with gaussian splatting},
  author={Gan, Wanshui and Liu, Fang and Xu, Hongbin and Mo, Ningkai and Yokoya, Naoto},
  booktitle={2025 IEEE/CVF International Conference on Computer Vision (ICCV)},
  pages={28980--28990},
  year={2025},
  organization={IEEE}
}

@inproceedings{wei2023surroundocc,
  title={Surroundocc: Multi-camera 3d occupancy prediction for autonomous driving},
  author={Wei, Yi and Zhao, Linqing and Zheng, Wenzhao and Zhu, Zheng and Zhou, Jie and Lu, Jiwen},
  booktitle={2023 IEEE/CVF International Conference on Computer Vision (ICCV)},
  pages={21672--21683},
  year={2023},
  organization={IEEE}
}

@inproceedings{huang2024selfocc,
  title={Selfocc: Self-supervised vision-based 3d occupancy prediction},
  author={Huang, Yuanhui and Zheng, Wenzhao and Zhang, Borui and Zhou, Jie and Lu, Jiwen},
  booktitle={2024 IEEE/CVF Conference on Computer Vision and Pattern Recognition (CVPR)},
  pages={19946--19956},
  year={2024},
  organization={IEEE}
}

@article{zhang2025occnerf,
  title={Occnerf: Advancing 3d occupancy prediction in lidar-free environments},
  author={Zhang, Chubin and Yan, Juncheng and Wei, Yi and Li, Jiaxin and Liu, Li and Tang, Yansong and Duan, Yueqi and Lu, Jiwen},
  journal={IEEE Transactions on Image Processing},
  volume={34},
  pages={3096--3107},
  year={2025},
  publisher={IEEE}
}

@inproceedings{yu2025language,
  title={Language driven occupancy prediction},
  author={Yu, Zhu and Pang, Bowen and Liu, Lizhe and Zhang, Runmin and Li, Qiang and Cao, Si-Yuan and Luo, Maochun and Chen, Mingxia and Yang, Sheng and Shen, Hui-Liang},
  booktitle={2025 IEEE/CVF International Conference on Computer Vision (ICCV)},
  pages={7548--7558},
  year={2025},
  organization={IEEE}
}

@inproceedings{lilja2026queryocc,
  title={QueryOcc: Query-based self-supervision for 3D semantic occupancy},
  author={Lilja, Adam and Lan, Ji and Fu, Junsheng and Hammarstrand, Lars},
  booktitle={Proceedings of the IEEE/CVF Conference on Computer Vision and Pattern Recognition},
  pages={21399--21408},
  year={2026}
}

@article{yu2023flashocc,
  title={Flashocc: Fast and memory-efficient occupancy prediction via channel-to-height plugin},
  author={Yu, Zichen and Shu, Changyong and Deng, Jiajun and Lu, Kangjie and Liu, Zongdai and Yu, Jiangyong and Yang, Dawei and Li, Hui and Chen, Yan},
  journal={arXiv preprint arXiv:2311.12058},
  year={2023}
}

@inproceedings{liu2024fully,
  title={Fully sparse 3d occupancy prediction},
  author={Liu, Haisong and Chen, Yang and Wang, Haiguang and Yang, Zetong and Li, Tianyu and Zeng, Jia and Chen, Li and Li, Hongyang and Wang, Limin},
  booktitle={European Conference on Computer Vision},
  pages={54--71},
  year={2024},
  organization={Springer}
}

@inproceedings{ma2024cotr,
  title={Cotr: Compact occupancy transformer for vision-based 3d occupancy prediction},
  author={Ma, Qihang and Tan, Xin and Qu, Yanyun and Ma, Lizhuang and Zhang, Zhizhong and Xie, Yuan},
  booktitle={2024 IEEE/CVF Conference on Computer Vision and Pattern Recognition (CVPR)},
  pages={19936--19945},
  year={2024},
  organization={IEEE}
}

@article{wang2024opus,
  title={Opus: occupancy prediction using a sparse set},
  author={Wang, Jiabao and Liu, Zhaojiang and Meng, Qiang and Yan, Liujiang and Wang, Ke and Yang, Jie and Liu, Wei and Hou, Qibin and Cheng, Ming-Ming},
  journal={Advances in Neural Information Processing Systems},
  volume={37},
  pages={119861--119885},
  year={2024}
}

@article{moon2025streaming,
  title={Streaming Dense Voxel Representations for 3D Occupancy Prediction},
  author={Moon, Seokha and Baek, Janghyun and Jeong, Yujin and Chae, Daewon and Kim, Giseop and Lee, Jungbeom and Kim, Jinkyu and Choi, Sunwook},
  journal={arXiv preprint arXiv:2503.22087},
  year={2025}
}

@inproceedings{chen2025alocc,
  title={Alocc: Adaptive lifting-based 3d semantic occupancy and cost volume-based flow predictions},
  author={Chen, Dubing and Fang, Jin and Han, Wencheng and Cheng, Xinjing and Yin, Junbo and Xu, Chengzhong and Khan, Fahad Shahbaz and Shen, Jianbing},
  booktitle={Proceedings of the IEEE/CVF International Conference on Computer Vision},
  pages={4156--4166},
  year={2025}
}

@inproceedings{li2024viewformer,
  title={Viewformer: Exploring spatiotemporal modeling for multi-view 3d occupancy perception via view-guided transformers},
  author={Li, Jinke and He, Xiao and Zhou, Chonghua and Cheng, Xiaoqiang and Wen, Yang and Zhang, Dan},
  booktitle={European Conference on Computer Vision},
  pages={90--106},
  year={2024},
  organization={Springer}
}

@inproceedings{hou2024fastocc,
  title={Fastocc: Accelerating 3d occupancy prediction by fusing the 2d bird’s-eye view and perspective view},
  author={Hou, Jiawei and Li, Xiaoyan and Guan, Wenhao and Zhang, Gang and Feng, Di and Du, Yuheng and Xue, Xiangyang and Pu, Jian},
  booktitle={2024 IEEE International Conference on Robotics and Automation (ICRA)},
  pages={16425--16431},
  year={2024},
  organization={IEEE}
}

@inproceedings{philion2020lift,
  title={Lift, splat, shoot: Encoding images from arbitrary camera rigs by implicitly unprojecting to 3d},
  author={Philion, Jonah and Fidler, Sanja},
  booktitle={European conference on computer vision},
  pages={194--210},
  year={2020},
  organization={Springer}
}

@article{li2024bevformer,
  title={Bevformer: learning bird’s-eye-view representation from lidar-camera via spatiotemporal transformers},
  author={Li, Zhiqi and Wang, Wenhai and Li, Hongyang and Xie, Enze and Sima, Chonghao and Lu, Tong and Yu, Qiao and Dai, Jifeng},
  journal={IEEE Transactions on Pattern Analysis and Machine Intelligence},
  volume={47},
  number={3},
  pages={2020--2036},
  year={2024},
  publisher={IEEE}
}

@article{zhu2026dr,
  title={Dr. Occ: Depth-and Region-Guided 3D Occupancy from Surround-View Cameras for Autonomous Driving},
  author={Zhu, Xubo and Zhang, Haoyang and He, Fei and Wu, Rui and Shan, Yanhu and Yang, Wen and Yu, Huai},
  journal={arXiv preprint arXiv:2603.01007},
  year={2026}
}

@inproceedings{liao2025stcocc,
  title={Stcocc: Sparse spatial-temporal cascade renovation for 3d occupancy and scene flow prediction},
  author={Liao, Zhimin and Wei, Ping and Chen, Shuaijia and Wang, Haoxuan and Ren, Ziyang},
  booktitle={2025 IEEE/CVF Conference on Computer Vision and Pattern Recognition (CVPR)},
  pages={1516--1526},
  year={2025},
  organization={IEEE}
}

@article{li2023fb,
  title={Fb-occ: 3d occupancy prediction based on forward-backward view transformation},
  author={Li, Zhiqi and Yu, Zhiding and Austin, David and Fang, Mingsheng and Lan, Shiyi and Kautz, Jan and Alvarez, Jose M},
  journal={arXiv preprint arXiv:2307.01492},
  year={2023}
}

@article{wu2026height,
  title={Height-Guided Projection Reparameterization for Camera-LiDAR Occupancy},
  author={Wu, Yuan and Yan, Zhiqiang and Lian, Jiawei and Wang, Zhengxue and Yang, Jian},
  journal={arXiv preprint arXiv:2605.05072},
  year={2026}
}

@article{wang2026moge,
  title={Moge-2: Accurate monocular geometry with metric scale and sharp details},
  author={Wang, Ruicheng and Xu, Sicheng and Dong, Yue and Deng, Yu and Xiang, Jianfeng and Lv, Zelong and Sun, Guangzhong and Tong, Xin and Yang, Jiaolong},
  journal={Advances in Neural Information Processing Systems},
  volume={38},
  pages={35928--35959},
  year={2026}
}

@article{lin2025depth,
  title={Depth anything 3: Recovering the visual space from any views},
  author={Lin, Haotong and Chen, Sili and Liew, Junhao and Chen, Donny Y and Li, Zhenyu and Shi, Guang and Feng, Jiashi and Kang, Bingyi},
  journal={arXiv preprint arXiv:2511.10647},
  year={2025}
}

@article{piccinelli2025unidepthv2,
  title={Unidepthv2: Universal monocular metric depth estimation made simpler},
  author={Piccinelli, Luigi and Sakaridis, Christos and Yang, Yung-Hsu and Segu, Mattia and Li, Siyuan and Abbeloos, Wim and Van Gool, Luc},
  journal={IEEE Transactions on Pattern Analysis and Machine Intelligence},
  year={2025},
  publisher={IEEE}
}

@inproceedings{pan2024renderocc,
  title={Renderocc: Vision-centric 3d occupancy prediction with 2d rendering supervision},
  author={Pan, Mingjie and Liu, Jiaming and Zhang, Renrui and Huang, Peixiang and Li, Xiaoqi and Xie, Hongwei and Wang, Bing and Liu, Li and Zhang, Shanghang},
  booktitle={2024 IEEE International Conference on Robotics and Automation (ICRA)},
  pages={12404--12411},
  year={2024},
  organization={IEEE}
}

@article{tian2023occ3d,
  title={Occ3d: A large-scale 3d occupancy prediction benchmark for autonomous driving},
  author={Tian, Xiaoyu and Jiang, Tao and Yun, Longfei and Mao, Yucheng and Yang, Huitong and Wang, Yue and Wang, Yilun and Zhao, Hang},
  journal={Advances in neural information processing systems},
  volume={36},
  pages={64318--64330},
  year={2023}
}

@inproceedings{tong2023scene,
  title={Scene as occupancy},
  author={Tong, Wenwen and Sima, Chonghao and Wang, Tai and Chen, Li and Wu, Silei and Deng, Hanming and Gu, Yi and Lu, Lewei and Luo, Ping and Lin, Dahua and others},
  booktitle={2023 IEEE/CVF International Conference on Computer Vision (ICCV)},
  pages={8372--8381},
  year={2023},
  organization={IEEE}
}

@inproceedings{wang2023openoccupancy,
  title={Openoccupancy: A large scale benchmark for surrounding semantic occupancy perception},
  author={Wang, Xiaofeng and Zhu, Zheng and Xu, Wenbo and Zhang, Yunpeng and Wei, Yi and Chi, Xu and Ye, Yun and Du, Dalong and Lu, Jiwen and Wang, Xingang},
  booktitle={2023 IEEE/CVF International Conference on Computer Vision (ICCV)},
  pages={17804--17813},
  year={2023},
  organization={IEEE}
}

@article{huang2021bevdet,
  title={Bevdet: High-performance multi-camera 3d object detection in bird-eye-view},
  author={Huang, Junjie and Huang, Guan and Zhu, Zheng and Ye, Yun and Du, Dalong},
  journal={arXiv preprint arXiv:2112.11790},
  year={2021}
}

@inproceedings{wu2025deep,
  title={Deep height decoupling for precise vision-based 3d occupancy prediction},
  author={Wu, Yuan and Yan, Zhiqiang and Wang, Zhengxue and Li, Xiang and Hui, Le and Yang, Jian},
  booktitle={2025 IEEE International Conference on Robotics and Automation (ICRA)},
  pages={12647--12654},
  year={2025},
  organization={IEEE}
}

@inproceedings{liu2023sparsebev,
  title={Sparsebev: High-performance sparse 3d object detection from multi-camera videos},
  author={Liu, Haisong and Teng, Yao and Lu, Tao and Wang, Haiguang and Wang, Limin},
  booktitle={2023 IEEE/CVF International Conference on Computer Vision (ICCV)},
  pages={18534--18544},
  year={2023},
  organization={IEEE}
}

@inproceedings{shi2024occupancy,
  title={Occupancy as set of points},
  author={Shi, Yiang and Cheng, Tianheng and Zhang, Qian and Liu, Wenyu and Wang, Xinggang},
  booktitle={European Conference on Computer Vision},
  pages={72--87},
  year={2024},
  organization={Springer}
}

@article{chen2026fdr,
  title={FDR-Occ: Factorized Dense Routing for Full-Spectrum 3D Occupancy Prediction},
  author={Chen, Dubing and Zheng, Huan and Yan, Tianyi and Zhou, Yucheng and Tao, Runzhou and Qiu, Zhongying and Yang, Jianfei and Shen, Jianbing},
  journal={arXiv preprint arXiv:2607.03822},
  year={2026}
}

@inproceedings{wang2024panoocc,
  title={Panoocc: Unified occupancy representation for camera-based 3d panoptic segmentation},
  author={Wang, Yuqi and Chen, Yuntao and Liao, Xingyu and Fan, Lue and Zhang, Zhaoxiang},
  booktitle={2024 IEEE/CVF Conference on Computer Vision and Pattern Recognition (CVPR)},
  pages={17158--17168},
  year={2024},
  organization={IEEE}
}

@inproceedings{chen2025sa,
  title={Sa-occ: Satellite-assisted 3d occupancy prediction in real world},
  author={Chen, Chen and Wang, Zhirui and Sheng, Taowei and Jiang, Yi and Li, Yundu and Cheng, Peirui and Zhang, Luning and Chen, Kaiqiang and Hu, Yanfeng and Yang, Xue and others},
  booktitle={2025 IEEE/CVF International Conference on Computer Vision (ICCV)},
  pages={27021--27030},
  year={2025},
  organization={IEEE}
}

@inproceedings{caesar2020nuscenes,
  title={nuscenes: A multimodal dataset for autonomous driving},
  author={Caesar, Holger and Bankiti, Varun and Lang, Alex H and Vora, Sourabh and Liong, Venice Erin and Xu, Qiang and Krishnan, Anush and Pan, Yu and Baldan, Giancarlo and Beijbom, Oscar},
  booktitle={2020 IEEE/CVF conference on computer vision and pattern recognition (CVPR)},
  pages={11618--11628},
  year={2020},
  organization={IEEE}
}

@inproceedings{he2016deep,
  title={Deep residual learning for image recognition},
  author={He, Kaiming and Zhang, Xiangyu and Ren, Shaoqing and Sun, Jian},
  booktitle={Proceedings of the IEEE conference on computer vision and pattern recognition},
  pages={770--778},
  year={2016}
}

@inproceedings{li2023bevdepth,
  title={Bevdepth: Acquisition of reliable depth for multi-view 3d object detection},
  author={Li, Yinhao and Ge, Zheng and Yu, Guanyi and Yang, Jinrong and Wang, Zengran and Shi, Yukang and Sun, Jianjian and Li, Zeming},
  booktitle={Proceedings of the AAAI conference on artificial intelligence},
  volume={37},
  number={2},
  pages={1477--1485},
  year={2023}
}

@inproceedings{li2023bevstereo,
  title={Bevstereo: Enhancing depth estimation in multi-view 3d object detection with temporal stereo},
  author={Li, Yinhao and Bao, Han and Ge, Zheng and Yang, Jinrong and Sun, Jianjian and Li, Zeming},
  booktitle={Proceedings of the AAAI conference on artificial intelligence},
  volume={37},
  number={2},
  pages={1486--1494},
  year={2023}
}

@article{loshchilov2017decoupled,
  title={Decoupled weight decay regularization},
  author={Loshchilov, Ilya and Hutter, Frank},
  journal={arXiv preprint arXiv:1711.05101},
  year={2017}
}

@article{chen2019mmdetection,
  title={MMDetection: Open mmlab detection toolbox and benchmark},
  author={Chen, Kai and Wang, Jiaqi and Pang, Jiangmiao and Cao, Yuhang and Xiong, Yu and Li, Xiaoxiao and Sun, Shuyang and Feng, Wansen and Liu, Ziwei and Xu, Jiarui and others},
  journal={arXiv preprint arXiv:1906.07155},
  year={2019}
}

@inproceedings{du2026sparseworld,
  title={Sparseworld-tc: Trajectory-conditioned sparse occupancy world model},
  author={Du, Jiayuan and Zhao, Yiming and Guo, Zhenglong and Pan, Yong and Hou, Wenbo and Hao, Zhihui and Zhan, Kun and Chen, Qijun},
  booktitle={Proceedings of the IEEE/CVF Conference on Computer Vision and Pattern Recognition},
  pages={7425--7434},
  year={2026}
}

@inproceedings{liu2026drivepi,
  title={Drivepi: Spatial-aware 4d mllm for unified autonomous driving understanding, perception, prediction and planning},
  author={Liu, Zhe and Huang, Runhui and Yang, Rui and Yan, Siming and Wang, Zining and Hou, Lu and Lin, Di and Bai, Xiang and Zhao, Hengshuang},
  booktitle={Proceedings of the IEEE/CVF Conference on Computer Vision and Pattern Recognition},
  pages={3688--3698},
  year={2026}
}

@inproceedings{yang2025driving,
  title={Driving in the occupancy world: Vision-centric 4d occupancy forecasting and planning via world models for autonomous driving},
  author={Yang, Yu and Mei, Jianbiao and Ma, Yukai and Du, Siliang and Chen, Wenqing and Qian, Yijie and Feng, Yuxiang and Liu, Yong},
  booktitle={Proceedings of the AAAI Conference on Artificial Intelligence},
  volume={39},
  number={9},
  pages={9327--9335},
  year={2025}
}

@inproceedings{zheng2024occworld,
  title={Occworld: Learning a 3d occupancy world model for autonomous driving},
  author={Zheng, Wenzhao and Chen, Weiliang and Huang, Yuanhui and Zhang, Borui and Duan, Yueqi and Lu, Jiwen},
  booktitle={European conference on computer vision},
  pages={55--72},
  year={2024},
  organization={Springer}
}

@article{yu2024panoptic,
  title={Panoptic-flashocc: An efficient baseline to marry semantic occupancy with panoptic via instance center},
  author={Yu, Zichen and Shu, Changyong and Sun, Qianpu and Bian, Yifan and Wei, Xiaobao and Yu, Jiangyong and Liu, Zongdai and Yang, Dawei and Li, Hui and Chen, Yan},
  journal={arXiv preprint arXiv:2406.10527},
  year={2024}
}

@inproceedings{cao2026occany,
  title={Occany: Generalized unconstrained urban 3d occupancy},
  author={Cao, Anh-Quan and Vu, Tuan-Hung},
  booktitle={Conference on Computer Vision and Pattern Recognition (CVPR)},
  year={2026}
}

@article{liu2026occanyscene,
  title={OccAnyScene: Towards Unified Indoor-Outdoor 3D Occupancy Prediction},
  author={Liu, Junjie and Gan, Wanshui and Dai, Zitong and Cao, Guiping and Li, Yan and Chen, Ke and Jiang, Dongmei and Zhang, Jianguo and Lan, Xiangyuan},
  journal={arXiv preprint arXiv:2608.08696},
  year={2026}
}

@article{lian2026liauto,
  title={LiAuto-GeoX: Efficient Grounded Driving Transformer},
  author={Lian, Jiawei and Sun, Haoyi and Wu, Yang and Mu, Lifu and Wang, Siyuan and Hui, Le and Mao, Ning and Wei, Tao and Zhou, Pan and Zhan, Kun and others},
  journal={arXiv preprint arXiv:2606.05774},
  year={2026}
}

@article{wu2026see,
  title={See through the dark: Learning illumination-affined representations for nighttime occupancy prediction},
  author={Wu, Yuan and Yan, Zhiqiang and Zhang, Yigong and Li, Xiang and Yang, Jian},
  journal={Advances in Neural Information Processing Systems},
  volume={38},
  pages={67209--67233},
  year={2026}
}

@inproceedings{park2026s2go,
  title={S2go: Streaming sparse gaussian occupancy},
  author={Park, Jinhyung and Peng, Chensheng and Zheng, Wenzhao and Kitani, Kris and Zhan, Wei and others},
  booktitle={International Conference on Learning Representations},
  volume={2026},
  pages={62911--62932},
  year={2026}
}

@inproceedings{li2023voxformer,
  title={Voxformer: Sparse voxel transformer for camera-based 3d semantic scene completion},
  author={Li, Yiming and Yu, Zhiding and Choy, Christopher and Xiao, Chaowei and Alvarez, Jose M and Fidler, Sanja and Feng, Chen and Anandkumar, Anima},
  booktitle={2023 IEEE/CVF Conference on Computer Vision and Pattern Recognition (CVPR)},
  pages={9087--9098},
  year={2023},
  organization={IEEE}
}

@article{gao2026deformable,
  title={Deformable gaussian occupancy: Decoupling rigid and nonrigid motion with factorized distillation},
  author={Gao, Yang and Li, Wuyang and Luan, Po-Chien and Alahi, Alexandre},
  journal={arXiv preprint arXiv:2605.28587},
  year={2026}
}

@inproceedings{ji2025ocrfdet,
  title={Ocrfdet: object-centric radiance fields for multi-view 3d object detection in autonomous driving},
  author={Ji, Mingqian and Zhang, Shanshan and Yang, Jian},
  booktitle={2025 IEEE/CVF International Conference on Computer Vision (ICCV)},
  pages={24933--24942},
  year={2025},
  organization={IEEE}
}

@article{ji2026depthfusion,
  title={Depthfusion: Depth-aware hybrid feature fusion for LiDAR-camera 3D object detection},
  author={Ji, Mingqian and Zhang, Shanshan and Yang, Jian},
  journal={IEEE Transactions on Multimedia},
  year={2026},
  publisher={IEEE}
}

@article{zhou2026generalizing,
  title={Generalizing visual geometry priors to sparse gaussian occupancy prediction},
  author={Zhou, Changqing and Luo, Yueru and Chen, Changhao},
  journal={arXiv preprint arXiv:2602.21552},
  year={2026}
}
}

%     \small
%     \bibliographystyle{ieeenat_fullname}
%     \bibliography{main}
% }
\end{document}